\documentclass[letterpaper]{article} 
\usepackage{aaai2026}  
\usepackage{times}  
\usepackage{helvet}  
\usepackage{courier}  
\usepackage[hyphens]{url}  
\usepackage{graphicx} 
\usepackage{natbib}  
\usepackage{caption} 
\usepackage{algorithm}
\usepackage{algorithmic}
\usepackage{tabularx}
\usepackage{dirtytalk}
\usepackage{quoting}
\usepackage{subcaption}
\usepackage{amsmath}
\usepackage{url}
\usepackage{newfloat}
\usepackage{listings}
\usepackage{tabularx}
\usepackage{booktabs}
\usepackage{cleveref}
\usepackage{bbding}
\usepackage{multirow}

\quotingsetup{font=itshape,vskip=1pt, leftmargin=15pt}
\usepackage[most]{tcolorbox}

\newtcolorbox{mybox}{
enhanced,
boxrule=0pt,frame hidden,
borderline west={4pt}{0pt}{white!75!black},
colback=white,
sharp corners
}

\DeclareCaptionStyle{ruled}{labelfont=normalfont,labelsep=colon,strut=off} 
\floatstyle{ruled}
\newfloat{listing}{tb}{lst}{}
\floatname{listing}{Listing}

\title{Promoting Sustainable Web Agents: Benchmarking and Estimating Energy Consumption Through Empirical and Theoretical Analysis}

\author {
    Lars Krupp\textsuperscript{\rm 1, 2},
    Daniel Geißler\textsuperscript{\rm 1, 2},
    Vishal Banwari\textsuperscript{\rm 1, 2},
    Paul Lukowicz\textsuperscript{\rm 1, 2},
    Jakob Karolus\textsuperscript{\rm 1, 2}
}
\affiliations {
    \textsuperscript{\rm 1}RPTU Kaiserslautern-Landau\\
    \textsuperscript{\rm 2}German Research Center for Artificial Intelligence (DFKI)\\
    lars.krupp@dfki.de
}

\usepackage{bibentry}

\begin{document}

\newcommand{\samplemean}{\bar{x}}
\newcommand{\samplesd}{s}
\newcommand{\chatbot}{\emph{WAgent}}
\newcommand{\website}{\emph{Website}}
\newcommand{\studyOne}{\emph{Study 1}}
\newcommand{\studyTwo}{\emph{Study 2}}
\newcommand{\compcostlasermindact}{10}
\newcommand{\compmindactnmindact}{7}

\newcommand{\todol}[1]{\textsf{\textbf{\textcolor{green!55!blue}{[Lars: #1]}}}}
\newcommand{\todoj}[1]{\textsf{\textbf{\textcolor{yellow!55!red}{[Jakob: #1]}}}}
\newcommand{\todod}[1]{\textsf{\textbf{\textcolor{yellow!55!green}{[Daniel: #1]}}}}

\maketitle

\begin{abstract}

Web agents, like OpenAI's Operator and Google's Project Mariner, are powerful agentic systems pushing the boundaries of Large Language Models (LLM). They can autonomously interact with the internet at the user's behest, such as navigating websites, filling search masks, and comparing price lists.
Though web agent research is thriving, induced sustainability issues remain largely unexplored.
To highlight the urgency of this issue, we provide an initial exploration of the energy and $CO_2$ cost associated with web agents from both a theoretical ---via estimation--- and an empirical perspective ---by benchmarking. Our results show how different philosophies in web agent creation can severely impact the associated expended energy, and that more energy consumed does not necessarily equate to better results. We highlight a lack of transparency regarding disclosing model parameters and processes used for some web agents as a limiting factor when estimating energy consumption. 
Our work contributes towards a change in thinking of how we evaluate web agents, advocating for dedicated metrics measuring energy consumption in benchmarks. 
\end{abstract}

\begin{links}
    \link{Code}{https://github.com/DFKIEI/WebAgentEnergy}
\end{links}

\section{Introduction}

Web agents powered by large language models (LLMs) represent the next significant milestone in how we interact with the internet~\cite{projectmariner,openaiIntroducingOperator2025}. These systems are able to "browse the web" and interact with online environments in a manner similar to human users. This concept holds immense potential to revolutionize internet usage, potentially even replacing traditional web browsers as the primary means of accessing information. However, the substantial computational costs~\cite{Samsi2023From} of these systems still remain a significant challenge.

There is an ongoing discussion on the sustainability and environmental impact of developing and deploying LLMs~\cite{bender2021dangers}, which are core components in any web agent. OpenAI's GPT-3 has 175 billion parameters and is trained on 570 GB of data~\cite{brown2020language}, using tremendous amounts of resources for training and inference. This necessitates the creation of large-scale data centers with substantial energy consumption. 


Companies are addressing this challenge in strikingly different ways. Some have opted to simply expand their energy resources. Google, for instance, is investing in the construction of nuclear power plants to support its data centers~\cite{silva2024googlenuclear} --- a controversial and arguably unsustainable approach. Others are taking a more forward-thinking route by introducing reporting standards for the energy consumption across the lifecycle of LLMs, cf.~Mistral~\cite{mistral2024envaistandard}. These initiatives aim to promote transparency, encourage sustainable development, and incentivize more energy-efficient implementations.


LLM-heavy applications like web agents are among the most computationally intensive AI systems. Yet, this reality is largely invisible to end users. Tools such as OpenAI's Operator or Google’s Project Mariner present themselves as simple input fields. Indistinguishable --- on the surface --- from a search bar or any other interface powered by a commodity LLM. There is no immediate feedback to users about the energy consumption or environmental impact of their queries. The cost remains abstract, detached from the interaction itself. In the background, however, the infrastructure required to support such systems includes high-emission data centers. 
As more and more users adopt web agents, their cumulative energy impact will become significant. There is a pressing need to bring this aspect to the forefront and to assess web agents not just on their performance, but also on their energy efficiency, rewarding efficient implementations. This work highlights this critical gap by quantifying the energy consumption and $CO_2$ emissions of six web agents through both theoretical means and an empirical evaluation.

In our empirical evaluation, we directly benchmark the energy consumption of five web agents that use open-source LLMs on eight different GPUs using the Mind2Web benchmark~\cite{deng2024mind2web}. Our procedure allows for fast, simple and efficient energy benchmarking for any web agent using open-source LLMs and directly compares the web agents by their real energy consumption compared to their performance results. Our results show that more energy consumed does not equate to better results. The most energy efficient web agent, AutoWebGLM~\cite{lai2024autowebglm}, also performed best in terms of average step success rate (SSR), a popular metric for gauging web agent performance.



Our theoretical approach proposes a method to estimate the energy consumption web agents using proprietary LLMs. For such agents, benchmarking the energy consumption is not possible, forcing us to rely on information present in literature. We applied this estimation approach to LASER~\cite{ma2023laser}, a web agent using the proprietary LLM GPT-4 and MindAct~\cite{deng2024mind2web} as a counterexample. MindAct is a web agent using open-source LLMs and smart preprocessing for computational efficiency. Additionally, we benchmarked MindAct, acquiring its exact energy costs, thus allowing us to compare it to its theoretical estimation.
By estimating and comparing the amount of energy consumed by these web agents with vastly different design philosophies, we highlight the impact of a web agent's design on its energy consumption. Using a conservative estimation, LASER spends approximately \compcostlasermindact{} times more energy than MindAct.


To quantify the effectiveness of estimating web agent energy consumption we evaluated MindAct~\cite{deng2024mind2web} both using benchmarking and estimation. Our results shine a light on the impact that uncertainty has on the estimation of this metric. Even for MindAct, a fully open-source agent, we overestimate its energy consumption by a factor of \compmindactnmindact. We conclude that the estimation of web agent energy consumption should only be considered when benchmarking is not possible.




Our work encourages a change in thinking of how we evaluate web agent performance. We show the stark differences in energy consumption between web agents and propose metrics to report (for proprietary LLM-driven agents) and a benchmark to use (for open-source LLM-driven agents) which allow comparison of their energy efficiency. We advocate for a holistic evaluation of web agents incorporating dedicated metrics for energy consumption.

\section{Related Work}
With the rapid improvements of LLMs in recent years and their ever improving capabilities in tool-use~\cite{dubey2024llama} a new frontier of research has become possible. With the goal of building agents that can interact with the internet much like a human would, web agents recently are gaining traction~\cite{deng2024mind2web, yao2022webshop}. Approaches in web agent construction show a high diversity, varying in input modalities, processing steps, and employed LLM. Some only use HTML for their input~\cite{ma2023laser, deng2024mind2web}, others use the accessibility tree instead~\cite{dechezelles2024browsergymecosystemwebagent}, supplement it with screenshots~\cite{zheng2024gpt} or use screenshots exclusively, like Pix2Act~\cite{lu2024weblinx}. Further characteristics include preprocessing~\cite{deng2024mind2web,gur2023real}, and the integration of memory modules~\cite{ma2023laser}. While some agents use open-source models~\cite{deng2024mind2web,gur2023real}, most use proprietary models~\cite{ma2023laser, zheng2024gpt,yang2024agentoccamsimplestrongbaseline,zhang2024webpilotversatileautonomousmultiagent}.

Likewise, many different benchmarks to evaluate their performances are being proposed in rapid succession~\cite{yao2022webshop,deng2024mind2web,he2024webvoyagerbuildingendtoendweb}. While efforts to unify these benchmarks exist~\cite{dechezelles2024browsergymecosystemwebagent}, comparing the performance of different web agents is challenging. Additionally, no benchmark yet takes the energy consumption of web agents into consideration and penalizes inefficient agents.



Since modern web agents are driven by LLMs, contextualizing their growing emissions is vital.
In the earlier stages of LLM development, research estimated the $CO_2$ emissions of, at the time state-of-the-art, transformer models such as BERT and GPT-3.
For BERT~\cite{devlin2018bert}, trained in the US, Devlin et al. calculated a potential environmental impact of 0.754 metric tons of $CO_2$ for a single training of 79h on 64 Tesla V100 GPUs with an average utilization of 62.7\%.
For GPT-3, the predecessor of ChatGPT, it is estimated that around 550 metric tons of $CO_2$ emissions were produced to complete the full training, based on the US American energy mix, tremendously exceeding the previous estimations from BERT due to increased complexity and dataset size~\cite{shi2023thinking}. 
On top, a significant amount of energy is commonly wasted on ineffective versions of the LLM and for tuning the hyperparameter spaces~\cite{Verdecchia2023A}.

After training, the deployment of LLMs to the public introduces an additional and significant layer of environmental impact. According to Samsi et al.~\cite{Samsi2023From}, the energy demand during inference is influenced by several unpredictable factors, including user load and deployment duration. They propose \emph{energy per token} as a useful metric for evaluating the trade-off between performance and sustainability, particularly for quantized LLMs.
Additionally, the environmental impact of LLMs extends beyond energy consumption to include resource use, such as the water needed for cooling data centers, and the e-waste generated by the disposal of outdated hardware, highlighting the complexity and difficulty of calculating and comparing the LLM carbon footprint throughout the whole life cycle~\cite{patterson2021carbon}.

To support the development of more sustainable web agents, our work focuses specifically on the energy costs incurred during inference, addressing a critical component of their broader impact on the environment. We benchmark the energy consumption and carbon footprint of web agents and demonstrate that, due to lack of transparency, estimating the energy usage of proprietary LLM-driven agents is unreliable.

\section{Energy Consumptions of Web Agents}
Sustainability is becoming an increasingly pressing concern in the development and deployment of generative AI systems, particularly as web agents gain broader adoption. In this work, we aim to provide insights into the energy efficiency of several prominent web agents introduced in recent research. We solely focus on the energy efficiency during inference as it will outweigh fine-tuning cost with continued use and widespread use of web agents. Our approach is twofold. First, we conduct a practical evaluation of five web agents, all of which rely on open-source LLMs and are thus available for benchmarking, reflecting realistic deployment scenarios. Second, we present a theoretical estimation of energy consumption for one of the benchmarked agents and an additional agent that uses a proprietary model, where benchmarking is infeasible due to the lack of open-source access. By including a benchmarked agent in the estimation process, we are able to assess the accuracy of our theoretical estimation and demonstrate its applicability in case empirical measurements are not possible. \Cref{tab:agents} lists all web agents that we evaluated in this paper. We note that completely closed-source agents such as OpenAI's Operator~\cite{openaiIntroducingOperator2025} and Google's Project Mariner~\cite{projectmariner} are impossible to estimate as no details of their implementation are available.


\begin{table}[ht]
    \centering
    \begin{tabular}{llcc}
\toprule
\textbf{Web Agent} & \textbf{LLM} & \textbf{BM} & \textbf{Est.} \\
\midrule
\multicolumn{4}{l}{Fully open-source web agents}\\
\midrule
AutoWebGLM & ChatGLM3-6B & \Checkmark & \\
~\cite{lai2024autowebglm} & & &\\
MindAct & DeBERTa-86M,  & \Checkmark & \Checkmark\\
~\cite{deng2024mind2web} &flan-T5$_{XL}$ -2.85B&        &\\
MultiUI & UIX-Qwen2-8.03B & \Checkmark & \\
~\cite{liu2024harnessing} & & &\\
Synapse & CodeLlama-Instruct-7B & \Checkmark & \\
~\cite{zheng2023synapse} & & &\\
Synatra& CodeLlama-7B & \Checkmark & \\
~\cite{ou2024synatra} & & &\\
\midrule
\multicolumn{4}{l}{Open-source web agent with proprietary LLM}\\
\midrule
LASER & GPT-4 & & \Checkmark \\
~\cite{ma2023laser} & & &\\
\bottomrule
\end{tabular}
    \caption{Web agents evaluated in this paper. BM abbreviates benchmark. Est. abbreviates Estimate.}
    \label{tab:agents}
\end{table}

\subsection{Empirical Evaluation through Benchmarks}
The most precise way of measuring the sustainability of an open-source LLM-driven web agent is to benchmark its energy consumption. While benchmarking methods for the energy consumption of LLMs have been established~\cite{samsi2023words}, those methods are not sufficient in our case as they do not take into account how effectively the web agent navigates its environment. We chose the Mind2Web benchmark~\cite{deng2024mind2web} since it is easy to set up (no external server setup that has to be accessed), guarantees comparability (tasks are not changing, in contrast to live benchmarks), and is one of the most popular benchmarks when evaluating web agents~\cite{krupp2025quantifying}. In contrast to other web agent benchmarks~\cite{yao2022webshop,liu2018reinforcement}, Mind2Web consists of real-world websites. This allows for a realistic evaluation with respect to the average number of tokens per website. Mind2Web consists of 2350 tasks on 137 websites in 31 domains, with an average number of actions needed for task completion of 7.3 and an average of 1135 HTML elements per website.  The tasks are distributed over three splits: cross-domain (across task and environment generalization), cross-task (across task, same environment generalization), and cross-website (across website, same domain generalization).

We selected five popular web agents (see~\Cref{tab:agents}) that are fully open-source (including the LLM), reproducible and already utilize Mind2Web in their evaluation. 

\subsubsection{Setup}
We conducted the benchmarking on our cluster equipped with a variety of commonly employed GPUs for AI and Machine Learning, as detailed in \Cref{tab:gpu_comparison}. Additionally, we ran each web agent five times on each GPU to ensure stable results.
To retrieve the energy consumption of the GPUs, we modified the original web agent code to include the carbontracker library~\cite{anthony2020carbontracker}. By setting a start and end flag using this library, we can acquire the actual energy consumed by the executed code on each connected GPU.
%
    
\begin{table}[h]
    \centering    
    \begin{tabular}{llll}
\toprule
\textbf{GPU Model} & \textbf{Architecture} & \textbf{VRAM} & \textbf{FP32 (TFLOPS)}\\
\midrule
A100-SXM4 & Ampere & 40 GB& 19.5 \\
A100-PCIe & Ampere & 40 GB& 19.5 \\
RTX A6000 & Ampere & 48 GB& 38.7 \\
RTX 3090 & Ampere & 24 GB& 35.6 \\
H100-SXM5 & Hopper & 80 GB& 67 \\
H100-NVL & Hopper & 94 GB& 60 \\
H200-SXM5 & Hopper & 141 GB& 67 \\
L40S & Ada Lovelace & 48 GB& 91.61 \\
\bottomrule
\end{tabular}
    \caption{NVIDIA GPUs used in our setup, cf.~\cite{power2025gpuspecs}.}
    \label{tab:gpu_comparison}
\end{table}


\subsubsection{Results}
\begin{figure}[ht]
    \centering
    \includegraphics[width=\linewidth]{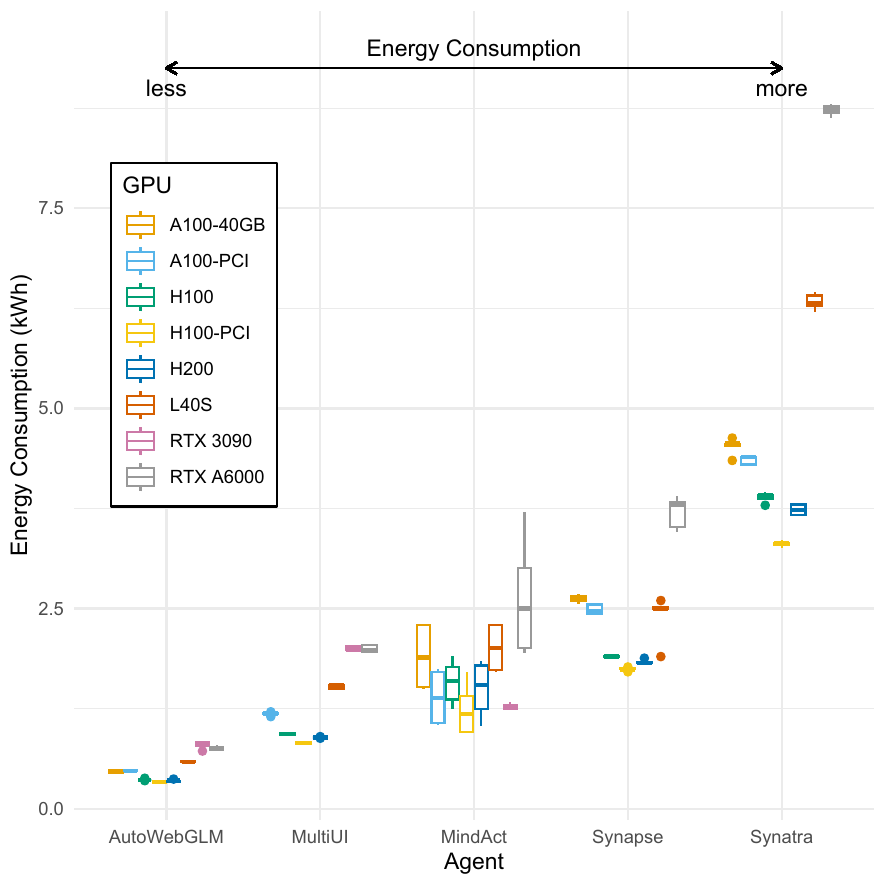}
    \caption{Energy consumption per web agent and GPU.}
    \label{fig:agent_energy_per_gpu}
\end{figure}

Our results show large differences in the energy consumption of different agents and GPUs. Clear trends become visible when analyzing \Cref{fig:agent_energy_per_gpu}, allowing for an ordering of web agents by energy consumption from most efficient to least efficient.

On average, the most energy-efficient GPU in our benchmarking tests was the Nvidia H100-NVL, a popular GPU for AI-related computing tasks. Hence, we choose to conduct further agent comparisons on this GPU only. 

We found large differences in energy consumption between the most efficient web agent and the most inefficient agent (see~\Cref{table:energy_to_SSR}). Synatra~\cite{ou2024synatra} consumes ten times more energy compared to AutoWebGLM~\cite{lai2024autowebglm}. However, the increased consumption does not lead to better results on the Mind2Web benchmark as reported by the agents' average step success rate (SSR). As such, AutoWebGLM is not only the most energy-efficient web agent, but it also performs best on the Mind2Web benchmark. 

The SSR is the de facto standard metric of the Mind2Web benchmark~\cite{krupp2025quantifying}, representing the ratio of successful steps towards the correct solution divided by the total steps. Since the Mind2Web benchmark is divided into three splits, each reporting a separate SSR, an average SSR~\cite{deng2024mind2web} is calculated at the end.

\begin{table*}[t]
    \centering
    \begin{tabular}{llrrr} 
        \toprule
        \textbf{Agent} & \textbf{Split} & \textbf{\# $10^6$ Tokens} & \textbf{Energy (kWh)} & \textbf{Energy/Token (kWh)} \\
        \midrule
        \multirow{3}{*}{AutoWebGLM}  & cross-domain & 0.25 & 0.22 $\pm$ 0.007 & $ (890 \pm 28.3) \times 10^{-9}$ \\
        & cross-task &0.09 & 0.06 $\pm$ 0.004 & $(719 \pm 46.42) \times 10^{-9}$ \\
        & cross-website & 0.06 & 0.05 $\pm$ 0.004 & $(831 \pm 69.28) \times 10^{-9}$ \\
        \addlinespace 
        \multirow{3}{*}{MindAct} & cross-domain & 183.78 & 0.66 $\pm$ 0.152 & $(3.59 \pm 0.08) \times 10^{-9}$ \\
        & cross-task & 78.27 & 0.35  $\pm$ 0.086 & $(4.47 \pm 1.10) \times 10^{-9}$ \\
        & cross-website & 49.60 & 0.21 $\pm$ 0.066  & $(4.23 \pm 1.33) \times 10^{-9}$ \\
        \addlinespace
        \multirow{3}{*}{MultiUI} & cross-domain & 1.59 & 0.52 $\pm$ 0.008 & $(326 \pm 5.04) \times 10^{-9}$ \\
        & cross-task & 0.65 & 0.18 & $(280 \pm 6.15) \times 10^{-9}$ \\
        & cross-website & 0.40 & 0.12 & $(307 \pm 10.06) \times 10^{-9}$ \\
        \addlinespace
        \multirow{3}{*}{Synapse} & cross-domain & 6.88 & 1.07 $\pm$ 0.018 & $(156\pm 2.62) \times 10^{-9}$ \\
        & cross-task & 2.97 & 0.42  $\pm$ 0.004 & $(142 \pm 1.35) \times 10^{-9}$ \\
        & cross-website & 1.93 & 0.25 $\pm$ 0.004 & $(131\pm 2.07) \times 10^{-9}$ \\
        \addlinespace
        \multirow{3}{*}{Synatra} & cross-domain & 24.34 & 2.11 $\pm$ 0.027 & $(86.7 \pm 1.11) \times 10^{-9}$ \\
        & cross-task & 8.92 & 0.72 $\pm$ 0.008 & $(80.9 \pm 0.9) \times 10^{-9}$ \\
        & cross-website & 5.50 & 0.48 $\pm$ 0.004 & $(86.9 \pm 0.73) \times 10^{-9}$ \\
        \bottomrule
    \end{tabular}
    \caption{Mean energy consumption per benchmark split and energy per input-token level for the H100-NVL GPU; the total number of tokens is dependent on the LLM's tokenizer.}
    \label{tab:energy_per_token}
\end{table*}


\begin{table}[ht]
\centering
    \begin{tabular}{lrrl}
        \toprule
        Agent & Ø SSR & Energy (kWh) & Time (min)  \\
        \midrule
        AutoWebGLM&\textbf{53.53}&\textbf{0.33} $\pm$ \textbf{0.01}&\textbf{\: 57.0} $\pm$ \textbf{0.8}\\
         MindAct & 43.50 & 1.22 $\pm$ 0.29 & 296.0 $\pm$ 90.2\\
         MultiUI & 34.70&0.82 $\pm$ 0.01 & 130.0 $\pm$ 1.2\\
         Synapse &21.67&1.74 $\pm$ 0.02& 356.0 $\pm$ 2.8\\
         Synatra&15.85&3.31 $\pm$ 0.04&426.0 $\pm$ 1.4\\
        \bottomrule
    \end{tabular}
    \caption{Energy consumption, computation time and reported average step success rate (SSR) per web agent on the Nvidia H100-NVL GPU.}
    \label{table:energy_to_SSR}
\end{table}

Within these splits, cross-domain contains the most tasks, followed by cross-task and cross-website. To account for the different sizes, we calculated the average energy spent per token for each web agent and for each split on the Nvidia H100-NVL GPU (see \Cref{tab:energy_per_token}). Our results show small differences between the splits for the same web agent. Additionally, it shows the importance of reducing the amount of ingested tokens into the LLM for the overall energy consumption. While the energy per token was consistently highest for AutoWebGLM, the overall energy consumed was lowest due to AutoWebGLM's preprocessing, which allowed the agent to significantly reduce the total amount of processed tokens. The energy per token metric is mostly influenced by LLM size, different tokenizers and different internal workings of the agents.

\subsection{Theoretical Estimation}
\label{sec:evaluation}
Many web agents rely on proprietary LLMs, preventing direct energy measurements through benchmarking. In this section, we propose an approach using theoretical estimation of their energy consumption based on available literature. As such, it is important to have detailed knowledge about the internal workings of each individual web agent to expose computationally inefficient implementations, making even this approach not feasible for completely closed-source agents such as OpenAI's Operator and Google's Project Mariner. However, if only the LLM is closed-source (proprietary), a theoretical estimation is possible if we can access information on the agent's internal procedure, e.g., through publications.



With MindAct~\cite{deng2024mind2web} and LASER~\cite{ma2023laser}, we chose two web agents differing in many aspects to provide an example of the diversity in approaches and mentalities present in the field. While MindAct uses comparatively small open-source models and does extensive preprocessing to get the best possible performance out of the available resources, LASER uses a proprietary model at its core with minimal preprocessing being done.
To estimate the energy consumption for each model, we analyzed and dissected available resources such as the accompanying publications~\cite{deng2024mind2web,ma2023laser} and the publicly available source code\footnote{MindAct: \url{https://github.com/OSU-NLP-Group/Mind2Web}, LASER: \url{https://github.com/Mayer123/LASER}}.


\subsubsection{MindAct}
\label{sec:mindact}
\begin{figure}[ht]
    \centering
    \includegraphics[width=\linewidth]{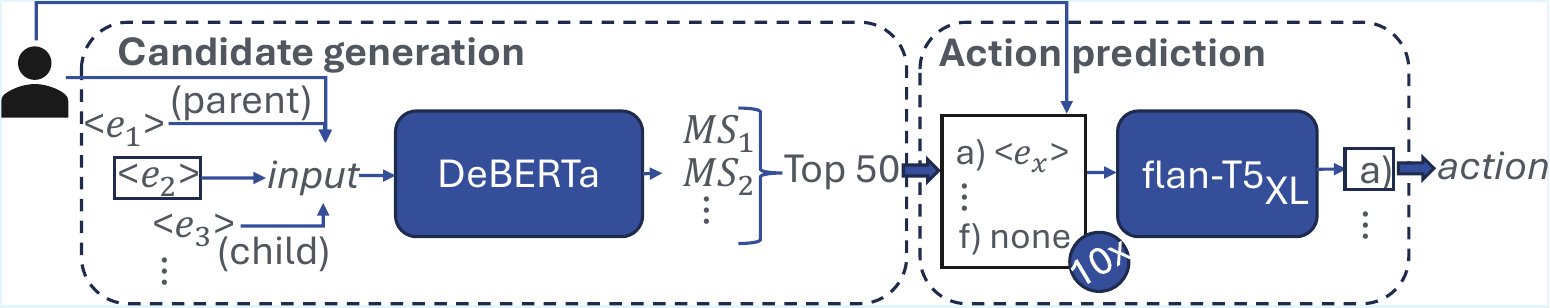}
    \caption{Pipeline depicting how an action is chosen in MindAct.}
    \label{fig:mindact_pipe}
\end{figure}

MindAct~\cite{deng2024mind2web} divides the process of finding the next correct \textit{action} to execute on the web into two stages, as depicted in \Cref{fig:mindact_pipe}. The first stage, candidate generation, is treated as a ranking task. Here, their finetuned DeBERTa 86M~\cite{he2021debertav3} model is given the following \textit{input}: the user's query, previous actions and a cleaned representation of each element in the Document Object Model (DOM) of the HTML webpage. Each cleaned element consists of its tag, its textual content, its salient attribute values and a textual representation of its respective parent and child nodes. From this, DeBERTa estimates a matching value $MS_i$ between 0 and 1 --- indicating how well an element matches the given user query. This process is repeated for all elements in the DOM. We estimate that the total number of tokens processed by DeBERTa at the end of this process is at least equivalent to the number of tokens in the original HTML. After processing all elements, the 50 elements with the highest matching values are used in the second stage. 
The second stage, action prediction, is constructed as a multiple-choice question answering challenge. A finetuned flan-T5$_{XL}$ model~\cite{chungScalingInstructionFinetunedLanguage2022} is given the user query, five of the returned elements and a none element and is tasked to decide which element is most likely to help towards answering the user query (the \textit{action} to perform). This is done a minimum of 10 times until all 50 returned elements are processed. The process is repeated if more than one possible action (besides none) is returned, until only one action remains (or all are rejected). For our estimation, we assume the maximum input length of flan-T5$_{XL}$~\cite{raffel2020exploring} (512 tokens) for one multiple-choice question and that a final result is obtained after the first pass (querying flan-T5$_{XL}$ 10 times).

To estimate the amount of energy per action for MindAct, we need to acquire (1) the energy per token for its LLMs, DeBERTa and flan-T5$_{XL}$, and (2) an estimation of the LLMs' context sizes (number of tokens passed to the LLM). Note that we already fixed $\overline{N}_{flan-T5_{XL}}=512$, the maximum input length of flan-T5$_{XL}$. For DeBERTa, we calculated the average number of tokens contained within an HTML page for the Mind2Web benchmark~\cite{deng2024mind2web} to be $\overline{N}_{DeBERTa}=118798$ using its tokenizer~\cite{he2021debertav3}. 
We extracted the energy per token for both models from our benchmark results (Nvidia H100-NVL), yielding $e_{DeBERTa}=3.77\cdot10^{-6}\,Wh$ and $e_{flan-T5_{XL}}=9.08\cdot 10^{-6}\,Wh$.
Resulting in the following energy per action for MindAct:
\scriptsize
\begin{align}
E_{action} &= E_{candidate\ generation} + E_{action\ prediction}\notag\\
E_{action} &= (\overline{N}_{DeBERTa} \cdot e_{DeBERTa})\notag \\
 & + 10 \cdot (\overline{N}_{flan-T5_{XL}} \cdot e_{flan-T5_{XL}})\notag\\
E_{action} &= \underline{\underline{0.49}}\,Wh\\
\notag
\end{align}
\normalsize
\noindent
Since the Mind2Web benchmark consists of 2350 tasks with an average of 7.3 actions per task, the total energy consumption $E_{total}$ equates to:
\scriptsize
\begin{align}
E_{total} &= E_{action} \cdot 7.3 \cdot 2350 = \underline{\underline{8.5}}\,kWh\label{actionMA_min}\\
\notag
\end{align}
\normalsize


\subsubsection{LASER}
\begin{figure}[t]
    \centering
    \includegraphics[width=\linewidth]{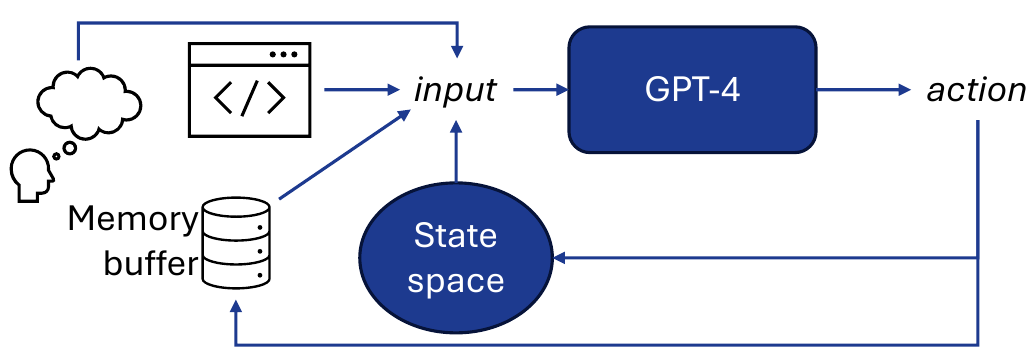}
    \caption{Pipeline depicting how an action is chosen in LASER.}
    \label{fig:laser_pipe}
\end{figure}

In contrast to MindAct, LASER~\cite{ma2023laser} makes use of a proprietary language model, specifically GPT-4~\cite{openai2024gpt4technicalreport}. LASER introduces states and state transitions for web agents (see~\Cref{fig:laser_pipe}), allowing for better recovery from mistakes and restricting the possible \textit{actions} depending on the state of the agent. LASER uses one-shot prompting and makes the model think step-by-step to improve the model's capabilities when dealing with complex user queries. Additionally, LASER has access to a memory buffer to store and access intermediary results (previous \textit{actions}). Finally, LASER is forced to produce a result after a maximum of 15 actions were generated.
However, the authors do not specify explicitly what the \textit{input} of their web agent is. We inferred that they use the raw, unmodified HTML by analyzing their results and LASER's benchmark: WebShop~\cite{yao2022webshop}. 

Analogously to MindAct's energy estimation, we need to acquire the energy per token for LASER's LLM, GPT-4, and an estimate for its context size.
We calculated the average number of tokens within an HTML page for the Mind2Web benchmark using the GPT-4~\cite{openai2024gpt4technicalreport} tokenizer at $\overline{N}_{GPT-4} = 93778$ tokens. Since GPT-4 is not open-source, we cannot execute the LLM and benchmark its energy consumption per token.

Consequently, estimating the energy consumption of GPT-4 on a per-token basis is inherently challenging due to a lack of publicly available technical specifications. The following estimation is based on leaked data and experts heuristics about OpenAI's GPT-4 model. Its exact size is unknown, though it is estimated to have 1.8 trillion parameters. Moreover, GPT-4 is reportedly using a mixture-of-experts (MoE) architecture comprised of 16 experts, each with around 111 billion parameters. However, in each forward pass only two of these are active at a time~\cite{schreiner2023gpt4architecture}, leading to about 222 billion active parameters.
The size of a model (the active parameters $N$) directly influences the number of floating-point operations (FLOP) required for inference, and thus its energy footprint. Using a conservative estimation for large parameter sizes (applicable for GPT-4)~\cite{kaplanScalingLawsNeural2020}, results in a compute cost for one forward-pass of $C_{forward}=2N$. Hence, roughly 444 billion FLOP are needed to compute one token using GPT-4.
Given a Nvidia H100 SXM GPU\footnote{Assuming H100 GPUs clusters~\cite{nvidiaNVIDIADGXH100} in OpenAI data centers.} with a theoretical maximum performance for FP8 Tensor Core of $2 \cdot 10^{15}\,$FLOP per second~\cite{NVIDIAH100GPU} in a dense configuration, yields a computation time for a single token on a H100 of about $2.22 \cdot 10^{-4}s$.

\begin{table*}[t] 
\centering
\begin{tabular}{llrrrr}
\toprule
\textbf{Method} &\textbf{Agent} & \textbf{Energy (kWh)} & &\textbf{Energy Mix}&\\
& & & Norway & US& Australia\\
 & & & 20g CO2e kWh & 453g CO2e kWh&800g CO2e kWh\\
\midrule
Benchmarking &AutoWebGLM & 0.33 & 6g& 149g & 264g\\
&MindAct &1.22 & 24g& 552g&976g\\
&MultiUI&0.82 & 16g& 371g&656g\\
&Synapse&1.74& 34g& 783g&1392g\\
&Synatra&3.31&66g&1499g&2648g\\
\midrule
Estimation & MindAct&9.01&180g&4081g&7208g\\
& LASER&99.21& 1984g& 44942g&79368g\\
 
\bottomrule
\end{tabular}
\caption{The $CO_2$ emissions of each web agent on the Mind2Web benchmark~\cite{deng2024mind2web} using the energy mixes of Norway, Australia~\cite{lannelongue2021green} and the US~\cite{co2us}.}
 \label{tab:co2}
\end{table*}

However, model size is only one component of the overall energy calculation. Other critical factors include GPU power draw~\cite{nvidiaNVIDIADGXH100}, actual utilization and power draw during inference~\cite{patelCharacterizingPowerManagement2024}, and data center overhead~\cite{butlerAWSGlobalData2024} (e.g., cooling, power distribution inefficiencies). In this estimation, we assume a best case scenario for GPU performance to provide a lower bound on energy costs. We note that the real energy costs are likely higher.

Given a DGX server of eight H100 GPUs with a maximum power rating of 10.2\,kW~\cite{nvidiaNVIDIADGXH100} and data center overheads of about 10-20\%~\cite{butlerAWSGlobalData2024} results in max 1.5\,kW per GPU. However, literature suggests GPU power consumption during inference is only about 70\%~\cite{patelCharacterizingPowerManagement2024}, leaving us with roughly 1\,kW power consumption per GPU during inference. Theoretically, this results in an estimated energy consumption per token of approximately $e_{GPT-4} = 0.22\,Ws = 6.17 \cdot 10^-5\,Wh$. We note that GPT-4 cannot be run on a single H100 GPU, but rather is run on a full cluster. This, however, does not change our calculation, since the computation time inversely scales with every additional GPU. As such, a DGX server with eight H100 will use eight times the power, but only requires one eighth of the computation time.

\scriptsize
\begin{align}
    E_{action} &= \overline{N}_{GPT-4} \cdot e_{GPT-4} = \underline{\underline{5.78}}Wh\\
    \notag
\end{align}

\normalsize
\noindent
Using Mind2Web task numbers as above, yields $E_{total}$:
\scriptsize
\begin{align}
    E_{total} &= E_{action} \cdot 7.3 \cdot 2350 = \underline{\underline{99.21}}\,kWh\\
\notag
\end{align}
\normalsize


\subsubsection{Comparing MindAct and LASER}
\label{sec:comparison}
Due to the lack of relevant information about the energy consumption of LLMs and the challenges involved in gathering reliable information as a third party, our results can only provide a conservative estimation. Using MindAct as an example --- estimated at $8.5\,kWh$ and benchmarked at $1.22\,kWh$ --- we showcase that a theoretical estimation approach works as a very coarse estimation in terms of orders of magnitude, highlighting the need for greater transparency and standardization in reporting the energy usage of LLMs. The best option still remains relying on actual benchmark results if available. However, our theoretical estimation allowed us valuable insights into the importance of adequate preprocessing and efficient implementations.

When comparing the energy consumption for LASER and MindAct, it becomes clear that the preprocessing done in MindAct is vital. Additionally, the smaller models, DeBERTa and flan-T5$_{XL}$, are much more energy efficient than the large GPT-4 model used in LASER.


\subsection{Carbon Dioxide Emissions}
To calculate the $CO_2$ emissions of the different web agents on the Mind2Web benchmark, we multiply their energy consumption by the $CO_2$ emissions per Wh. We provide results for multiple energy mixes in \Cref{tab:co2}. $CO_2$ emissions serve as a good proxy for energy consumption and provide end-users with a more tangible understanding of an agent's energy efficiency.


To further aid in understanding the differences between the agents, we additionally converted the $CO_2$ emissions into distance traveled by the average car, assuming 248.55\,g of $CO_2$ emissions per kilometer driven\footnote{\url{https://www.epa.gov/greenvehicles/greenhouse-gas-emissions-typical-passenger-vehicle#typical-passenger}}. The emissions generated by running AutoWebGLM on the Mind2Web benchmark equate to 0.6\,km traveled, assuming US emissions. For Synatra, the least efficient open-source LLM-driven agent, the range of the car equates to 6\,km. For running LASER once on the Mind2Web benchmark, one can drive up to 181\,km.

\section{Discussion}




While web agents are not widely used yet, it is increasingly evident that they will play a central role in how users interact with the internet. Examples like OpenAI’s Operator and Google’s Project Mariner already signal this paradigm shift. As these systems become more widespread and integrated into everyday web interactions, the importance of scrutinizing their environmental impact grows.

A critical problem is that the energy consumption of web agents is not readily apparent to the end user. Interfaces are often as simple as a search bar or text box, making it impossible for users to distinguish between “good” and “bad” agents, as their energy and environmental costs are hidden~\cite{mazzucatoUglyTruthChatGPT2024}. There's further value in communicating environmental costs directly to the user~\cite{li2023making}. Displaying estimated $CO_2$ emissions per task could help users become aware of the environmental impact of their interactions and guide them toward more sustainable agents. This not only encourages responsible development practices but also empowers users to make informed choices. However, vastly different energy mixes across countries (see~\Cref{tab:co2}) make comparison difficult and should be considered.

Our results highlight that the differences in energy consumption between web agents are quite extensive. Some agents leverage smart implementations and clever use of computational resources, achieving strong performance while consuming significantly less energy. Importantly, our findings show that energy efficiency does not compromise performance. In fact, some of the most efficient agents also perform well on benchmarks. Consequently, the choice of the internal model(s) is a crucial design decision that affects both energy efficiency and step success rate. Choosing a suitable LLM for one’s web agent that integrates well for its given task is an inherent trait of each web agent. This highlights a critical design choice in web agent development: the use of resource-intensive, brute-force approaches versus more efficient, fine-tuned implementations. Choosing the latter can result in significant environmental and computational advantages, as highlighted in our work.

However, incentives for developers are missing, as benchmarks do not penalize energy consumption---yet. To ensure sustainable use of web agents at scale, it is essential to evaluate them not only on performance, but also on energy efficiency. We advocate for augmenting existing benchmarks with standardized energy consumption metrics as we have done for Mind2Web\footnote{Source code available in our GitHub repo.}. We propose to use energy per benchmark as a core evaluation metric, enabling transparent comparison across systems.




\subsubsection{Benchmarking Open-Source Web Agents}

Our findings demonstrate that energy benchmarking does not require substantial overhead and can be conducted efficiently alongside standard performance evaluation. This enables a holistic assessment of web agents that captures both their task performance as well as energy efficiency. By incorporating energy consumption as a measurable and comparable metric, developers can identify implementations that achieve optimal task performance while minimizing energy consumption. This proposal supports informed design decisions and promotes the development of web agents that are not only powerful but also sustainable. As such, energy benchmarking should become a standard practice in the evaluation of open-source web agents, complementing accuracy-focused metrics such as step success rate with system-level efficiency.





\subsubsection{Estimating Energy Use of Proprietary Web Agents}
For web agents powered by proprietary LLMs, direct energy benchmarking is infeasible due to the lack of access to low-level system information, such as GPU usage. As a result, energy consumption must be estimated, which introduces significant uncertainty. Our comparison using MindAct, for which both estimation and direct benchmarking are possible, revealed discrepancies of up to a factor of \compmindactnmindact. This gap mainly stems from conservative, upper-bound token assumptions we made in MindAct’s candidate-generation stage. In practice, early termination, token truncation, and token reuse reduce the effective load, explaining the wide but unpredictable range between worst- and best-case energy usage and highlights how unreliable such estimations can be, even for a fully open-source agent.

Estimating energy use becomes even more problematic for proprietary LLMs, where energy per token cannot be measured and model parameters are undisclosed. If there is no other option, we recommend that developers report at the very least two key metrics: the energy consumption per token and, crucially, the number of tokens consumed. As we have shown in \Cref{tab:energy_per_token}, it is not enough for web agents to only report the energy consumption on a per token level, as recommended in previous literature~\cite{Samsi2023From}. Due to the structural complexity of web agents, often involving multiple LLM calls, preprocessing, and action steps, reporting energy per token alone is insufficient. For comparability, we propose to report the number of consumed tokens for established benchmarks, such as Mind2Web.

\section{Conclusion}
In this work, we compare the energy consumption of multiple web agents using both an empirical approach through benchmarking and a theoretical approach through estimation. We show that web agent design and used language models significantly influence the energy consumption and propose the introduction of web agent sustainability benchmarking to penalize inefficient energy consumption of web agents.

\section*{Acknowledgments}
This work is supported by the European Union’s Horizon Europe research and innovation program (HORIZON-CL4-2021-HUMAN-01) through the "SustainML" project (grant agreement No 101070408) and supported by the Carl Zeiss Foundation through the project "Sustainable Embedded AI".

\bibliography{aaai2026}
\onecolumn
\section{Supplementary Material}
In this supplementary material section, we provide detailed benchmark results for all web agents and all GPUs as benchmarked in our paper. These include the energy expendend for the whole benchmark (Mind2Web), as well as for individual splits of the benchmark, listing energy per token. We further extend our $CO_2$ calculations, providing a complete overview for all agents, all GPUs and the three energy mixes mentioned in our paper.

\subsection{Total Energy Consumption on Mind2Web Benchmark}
\Cref{tab:energy} shows the expended energy in kWh and time in minutes for all web agents on all GPUs on the Mind2Web benchmark~\cite{deng2024mind2web}, extending the table in the paper (only showing results for the H100-NVL).

\begin{table*}[h]
    \centering
    \begin{tabular}{llcc}
\toprule
Agent & GPU &  Energy(kWh) & Time (min)\\
\midrule
AutoWebGLM & A100-SXM4 & 0.46 $\pm$ 0.02 &  88.3  $\pm$ 3.3\\
AutoWebGLM& A100-PCIe  & 0.48 $\pm$ 0.01 &  88.4  $\pm$ 1.26\\
AutoWebGLM & H100-SXM5 & 0.36 $\pm$ 0.01 & 44.9  $\pm$ 1.06\\
AutoWebGLM  & H100-NVL & 0.33 $\pm$ 0.01 & 57.0  $\pm$ 0.8\\
AutoWebGLM & H200-SXM5 & 0.35 $\pm$ 0.02 &  44.5  $\pm$ 1.25\\
AutoWebGLM & L40S & 0.58 $\pm$ 0.01 &  81.1  $\pm$ 3.15\\
AutoWebGLM & RTX 3090 & 0.80 $\pm$ 0.05 &  91.6  $\pm$ 2.35\\
AutoWebGLM & RTX A6000 & 0.76 $\pm$ 0.02 & 103.0   $\pm$ 2.83\\
\addlinespace
 MindAct & A100-SXM4 & 1.90 $\pm$ 0.40 & 468.0  $\pm$ 92.6\\
 MindAct & A100-PCIe & 1.39 $\pm$ 0.34 & 459.0  $\pm$ 89.9\\
 MindAct & H100-SXM5 & 1.58 $\pm$ 0.26 & 326.0  $\pm$ 55.4\\
 MindAct & H100-NVL & 1.22 $\pm$ 0.29 & 301.9  $\pm$ 94.3\\
 MindAct & H200-SXM5 & 1.51 $\pm$ 0.31 & 300.0  $\pm$ 67.4\\
 MindAct & L40S & 2.01 $\pm$ 0.30 & 402.0  $\pm$ 55.3\\
 MindAct & RTX 3090 & 1.28 $\pm$ 0.04 & 202.0   $\pm$ 3.58\\
 MindAct & RTX A6000 & 2.57 $\pm$ 0.64 & 570.0 $\pm$ 126.0\\
\addlinespace
MultiUI & A100-PCIe & 1.18 $\pm$ 0.02 & 218.0   $\pm$ 1.79   \\
MultiUI & H100-SXM5 & 0.93 $\pm$ 0.02 & 113.0   $\pm$ 4.53   \\
MultiUI & H100-NVL & 0.822 $\pm$ 0.01 & 129.9   $\pm$ 1.20   \\
MultiUI & H200-SXM5 & 0.89 $\pm$ 0.01 & 107.0   $\pm$ 1.52   \\
MultiUI & L40S & 1.52 $\pm$ 0.03 & 209.0   $\pm$ 0.335  \\
MultiUI & RTX 3090 & 2.00 $\pm$ 0.03 & 223.0   $\pm$ 3.16   \\
MultiUI & RTX A6000 & 2.00 $\pm$ 0.05 & 269.0   $\pm$ 4.06   \\
 \addlinespace
 Synapse & A100-SXM4 & 2.62 $\pm$ 0.05 & 592.0  $\pm$ 5.00    \\
 Synapse & A100-PCIe & 2.48 $\pm$ 0.06 & 606.0  $\pm$ 13.9     \\
 Synapse & H100-SXM5 & 1.90 $\pm$ 0.01 & 277.0  $\pm$ 1.52    \\
 Synapse & H100-NVL & 1.74 $\pm$ 0.02 & 356.3  $\pm$ 2.80    \\
 Synapse & H200-SXM5 & 1.83 $\pm$ 0.03 & 276.0  $\pm$ 2.47    \\
 Synapse & L40S & 2.40 $\pm$ 0.28 & 339.0  $\pm$ 38.2    \\
 Synapse & RTX A6000 & 3.70 $\pm$ 0.20 & 574.0  $\pm$ 19.4     \\
 \addlinespace
 Synatra & A100-SXM4 & 4.53 $\pm$ 0.11 & 668.0    $\pm$ 2.27   \\
 Synatra & A100-PCIe & 4.36 $\pm$ 0.06 & 683.0    $\pm$ 2.15   \\
 Synatra & H100-SXM5 & 3.89 $\pm$ 0.06 & 409.0    $\pm$ 0.737  \\
 Synatra & H100-NVL & 3.31 $\pm$ 0.04 & 426.5    $\pm$ 1.40   \\
 Synatra & H200-SXM5 & 3.74 $\pm$ 0.07 & 369.0    $\pm$ 1.71   \\
 Synatra & L40S & 6.33 $\pm$ 0.10 & 906.0    $\pm$ 0.551  \\
 Synatra & RTX A6000 & 8.72 $\pm$ 0.07 & 1143.0    $\pm$ 2.04   \\
\bottomrule
\end{tabular}
    \caption{Energy and completion times per web agent and GPU}
    \label{tab:energy}
\end{table*}
\newpage

\subsection{Energy Consumption per Benchmark Split and Energy per Token}
We fruther compiled the detailed results for energy consumption per token for each web agent, the benchmark split and GPU. In \Cref{tab:AutoWebGLM}, the complete results for AutoWebGLM~\cite{lai2024autowebglm} are shown. \Cref{tab:MindAct} shows the complete results for MindAct~\cite{deng2024mind2web}, \Cref{tab:MultiUI} for MultiUI~\cite{liu2024harnessing}, \Cref{tab:Synatra} for Synatra~\cite{ou2024synatra}, and  \Cref{tab:synapse} for Synapse~\cite{zheng2023synapse}.

\begin{table*}[h]
    \centering
    \begin{tabular}{llrrr}
    \toprule
     Agent & GPU        & Split   &    Energy(kWh)  & Energy/Token ($10^{-9}$kWh) \\
     \midrule
AutoWebGLM & A100-SXM4  & domain & 0.308 $\pm$ 0.011 & 1245.35 $\pm$ 44.48\\
AutoWebGLM & A100-SXM4  & task & 0.088 $\pm$ 0.008 & 1021.18 $\pm$ 92.83\\
AutoWebGLM & A100-SXM4  & website & 0.066 $\pm$ 0.005 & 1143.15 $\pm$ 86.6\\
AutoWebGLM & A100-PCIe   & domain & 0.314 $\pm$ 0.009 & 1269.61 $\pm$ 36.39\\
AutoWebGLM & A100-PCIe   & task & 0.094 $\pm$ 0.005 & 1090.8 $\pm$ 58.02\\
AutoWebGLM & A100-PCIe   & website & 0.068 $\pm$ 0.004 & 1177.8 $\pm$ 69.28\\
AutoWebGLM & H100-SXM5       &   domain    & 0.242 $\pm$ 0.011 & 978.49 $\pm$ 44.48\\
AutoWebGLM & H100-SXM5       & task  & 0.072 $\pm$ 0.008 & 835.51 $\pm$ 92.83\\
AutoWebGLM & H100-SXM5       &website   & 0.048 $\pm$ 0.004 & 831.38 $\pm$ 69.28\\
AutoWebGLM & H100-NVL   & domain & 0.22 $\pm$ 0.007 & 889.54 $\pm$ 28.3\\
AutoWebGLM & H100-NVL   & task & 0.062 $\pm$ 0.004 & 719.47 $\pm$ 46.42\\
AutoWebGLM & H100-NVL   & website & 0.048 $\pm$ 0.004 & 831.38 $\pm$ 69.28\\
AutoWebGLM & H200-SXM5 & domain    & 0.232 $\pm$ 0.011 & 938.06 $\pm$ 44.48\\
AutoWebGLM & H200-SXM5 & task  & 0.066 $\pm$ 0.005 & 765.88 $\pm$ 58.02\\
AutoWebGLM & H200-SXM5 & website   & 0.05 $\pm$ 0.0 & 866.03 $\pm$ 0.0\\
AutoWebGLM & L40S & domain    & 0.388 $\pm$ 0.008 & 1568.82 $\pm$ 32.35\\
AutoWebGLM & L40S & task  & 0.112 $\pm$ 0.008 & 1299.68 $\pm$ 92.83\\
AutoWebGLM & L40S & website   & 0.084 $\pm$ 0.005 & 1454.92 $\pm$ 86.6\\
AutoWebGLM & RTX 3090   & domain & 0.532 $\pm$ 0.043 & 2151.06 $\pm$ 173.86\\
AutoWebGLM & RTX 3090   & task & 0.154 $\pm$ 0.005 & 1787.06 $\pm$ 58.02\\
AutoWebGLM & RTX 3090   & website & 0.114 $\pm$ 0.018 & 1974.54 $\pm$ 311.77\\
AutoWebGLM & RTX A6000 & domain    & 0.516 $\pm$ 0.026 & 2086.37 $\pm$ 105.13\\
AutoWebGLM & RTX A6000 & task  & 0.142 $\pm$ 0.008 & 1647.81 $\pm$ 92.83\\
AutoWebGLM & RTX A6000 & website   & 0.102 $\pm$ 0.004 & 1766.69 $\pm$ 69.28\\
\bottomrule
    \end{tabular}
    \caption{Energy cost per Token for each Split on all GPUs for the AutoWebGLM web agent}
    \label{tab:AutoWebGLM}
\end{table*}

\begin{table*}[]
    \centering
    \begin{tabular}{llrrr}
    \toprule
     Agent & GPU        & Split   &    Energy(kWh)  & Energy/Token ($10^{-9}$kWh) \\
     \midrule
MindAct    & A100-SXM4 & domain & 1.0 $\pm$ 0.162 & 6.63 $\pm$ 1.07\\
MindAct    & A100-SXM4 & task & 0.569 $\pm$ 0.145 & 8.41 $\pm$ 2.14\\
MindAct    & A100-SXM4 & website & 0.328 $\pm$ 0.097 & 7.78 $\pm$ 2.3\\
MindAct    & A100-PCIe & domain & 0.73 $\pm$ 0.132 & 4.84 $\pm$ 0.87\\
MindAct    & A100-PCIe & task & 0.418 $\pm$ 0.129 & 6.18 $\pm$ 1.91\\
MindAct    & A100-PCIe & website & 0.245 $\pm$ 0.083 & 5.81 $\pm$ 1.97\\
MindAct    & H100-SXM5 & domain & 0.823 $\pm$ 0.078 & 5.45 $\pm$ 0.52\\
MindAct    & H100-SXM5 & task & 0.483 $\pm$ 0.117 & 7.14 $\pm$ 1.73\\
MindAct    & H100-SXM5 & website & 0.271 $\pm$ 0.073 & 6.43 $\pm$ 1.73\\
MindAct    & H100-NVL & domain & 0.656 $\pm$ 0.152 & 4.47 $\pm$ 1.10\\
MindAct    & H100-NVL & task & 0.349 $\pm$ 0.086 & 4.23 $\pm$ 1.33\\
MindAct    & H100-NVL & website & 0.21 $\pm$ 0.066 & 4.98 $\pm$ 1.57\\
MindAct    & H200-SXM5 & domain & 0.806 $\pm$ 0.103 & 5.34 $\pm$ 0.68\\
MindAct    & H200-SXM5 & task & 0.454 $\pm$ 0.124 & 6.71 $\pm$ 1.83\\
MindAct    & H200-SXM5 & website & 0.274 $\pm$ 0.068 & 6.5 $\pm$ 1.61\\
MindAct    & L40S & domain & 1.09 $\pm$ 0.113 & 7.22 $\pm$ 0.75\\
MindAct    & L40S & task & 0.575 $\pm$ 0.115 & 8.5 $\pm$ 1.7\\
MindAct    & L40S & website & 0.341 $\pm$ 0.075 & 8.09 $\pm$ 1.78\\
MindAct    & RTX 3090 & domain & 0.47 $\pm$ 0.01 & 3.11 $\pm$ 0.07\\
MindAct    & RTX 3090 & task & 0.498 $\pm$ 0.024 & 7.36 $\pm$ 0.35\\
MindAct    & RTX 3090 & website & 0.308 $\pm$ 0.013 & 7.31 $\pm$ 0.31\\
MindAct    & RTX A6000 & domain & 1.4 $\pm$ 0.356 & 9.28 $\pm$ 2.36\\
MindAct    & RTX A6000 & task & 0.734 $\pm$ 0.206 & 10.85 $\pm$ 3.04\\
MindAct    & RTX A6000 & website & 0.428 $\pm$ 0.124 & 10.15 $\pm$ 2.94\\
\bottomrule
    \end{tabular}
    \caption{Energy cost per Token for each Split on all GPUs for the MindAct web agent}
    \label{tab:MindAct}
\end{table*}

\begin{table*}[]
    \centering
    \begin{tabular}{llrrr}
    \toprule
     Agent & GPU        & Split   &    Energy(kWh)  & Energy/Token ($10^{-9}$kWh) \\
     \midrule
MultiUI    & A100-PCIe & domain & 0.744 $\pm$ 0.011 & 468.49 $\pm$ 6.93\\
MultiUI    & A100-PCIe & task & 0.268 $\pm$ 0.008 & 412.35 $\pm$ 12.31\\
MultiUI    & A100-PCIe & website & 0.172 $\pm$ 0.008 & 432.53 $\pm$ 20.12\\
MultiUI    & H100-SXM5 & domain & 0.588 $\pm$ 0.022 & 370.26 $\pm$ 13.85\\
MultiUI    & H100-SXM5 & task & 0.206 $\pm$ 0.009 & 316.96 $\pm$ 13.85\\
MultiUI    & H100-SXM5 & website & 0.136 $\pm$ 0.005 & 342.0 $\pm$ 12.57\\
MultiUI    & H100-NVL & domain & 0.518 $\pm$ 0.008 & 326.18 $\pm$ 5.04\\
MultiUI    & H100-NVL & task & 0.182 $\pm$ 0.004 & 280.03 $\pm$ 6.15\\
MultiUI    & H100-NVL & website & 0.122 $\pm$ 0.004 & 306.8 $\pm$ 10.06\\
MultiUI    & H200-SXM5 & domain & 0.558 $\pm$ 0.008 & 351.37 $\pm$ 5.04\\
MultiUI    & H200-SXM5 & task & 0.2 $\pm$ 0.0 & 307.73 $\pm$ 0.0\\
MultiUI    & H200-SXM5 & website & 0.132 $\pm$ 0.004 & 331.94 $\pm$ 10.06\\
MultiUI    & L40S & domain & 0.966 $\pm$ 0.025 & 608.28 $\pm$ 15.74\\
MultiUI    & L40S & task & 0.336 $\pm$ 0.005 & 516.98 $\pm$ 7.69\\
MultiUI    & L40S & website & 0.222 $\pm$ 0.004 & 558.27 $\pm$ 10.06\\
MultiUI    & RTX 3090 & domain & 1.26 $\pm$ 0.019 & 793.41 $\pm$ 11.96\\
MultiUI    & RTX 3090 & task & 0.448 $\pm$ 0.016 & 689.31 $\pm$ 24.62\\
MultiUI    & RTX 3090 & website & 0.29 $\pm$ 0.017 & 729.27 $\pm$ 42.75\\
MultiUI    & RTX A6000 & domain & 1.24 $\pm$ 0.046 & 780.81 $\pm$ 28.97\\
MultiUI    & RTX A6000 & task & 0.46 $\pm$ 0.0 & 707.77 $\pm$ 0.0\\
MultiUI    & RTX A6000 & website & 0.3 $\pm$ 0.0 & 754.42 $\pm$ 0.0\\
\bottomrule
    \end{tabular}
    \caption{Energy cost per Token for each Split on all GPUs for the MultiUI web agent}
    \label{tab:MultiUI}
\end{table*}

\begin{table*}[]
    \centering
    \begin{tabular}{llrrr}
    \toprule
     Agent & GPU        & Split   &    Energy(kWh)  & Energy/Token ($10^{-9}$kWh) \\
     \midrule
Synatra    & A100-SXM4 & domain & 2.85 $\pm$ 0.101 & 117.11 $\pm$ 4.15\\
Synatra    & A100-SXM4 & task & 1.0 $\pm$ 0.015 & 112.05 $\pm$ 1.68\\
Synatra    & A100-SXM4 & website & 0.676 $\pm$ 0.009 & 122.9 $\pm$ 1.64\\
Synatra    & A100-PCIe & domain & 2.77 $\pm$ 0.052 & 113.82 $\pm$ 2.14\\
Synatra    & A100-PCIe & task & 0.946 $\pm$ 0.023 & 106.0 $\pm$ 2.58\\
Synatra    & A100-PCIe & website & 0.638 $\pm$ 0.004 & 115.99 $\pm$ 0.73\\
Synatra    & H100-SXM5 & domain & 2.48 $\pm$ 0.062 & 101.9 $\pm$ 2.55\\
Synatra    & H100-SXM5 & task & 0.852 $\pm$ 0.016 & 95.47 $\pm$ 1.79\\
Synatra    & H100-SXM5 & website & 0.558 $\pm$ 0.004 & 101.44 $\pm$ 0.73\\
Synatra    & H100-NVL & domain & 2.11 $\pm$ 0.027 & 86.7 $\pm$ 1.11\\
Synatra    & H100-NVL & task & 0.722 $\pm$ 0.008 & 80.9 $\pm$ 0.9\\
Synatra    & H100-NVL & website & 0.478 $\pm$ 0.004 & 86.9 $\pm$ 0.73\\
Synatra    & H200-SXM5 & domain & 2.38 $\pm$ 0.056 & 97.79 $\pm$ 2.3\\
Synatra    & H200-SXM5 & task & 0.814 $\pm$ 0.011 & 91.21 $\pm$ 1.23\\
Synatra    & H200-SXM5 & website & 0.54 $\pm$ 0.019 & 98.17 $\pm$ 3.45\\
Synatra    & L40S & domain & 4.02 $\pm$ 0.094 & 165.18 $\pm$ 3.86\\
Synatra    & L40S & task & 1.38 $\pm$ 0.053 & 154.64 $\pm$ 5.94\\
Synatra    & L40S & website & 0.932 $\pm$ 0.029 & 169.44 $\pm$ 5.27\\
Synatra    & RTX A6000 & domain & 5.61 $\pm$ 0.05 & 230.51 $\pm$ 2.05\\
Synatra    & RTX A6000 & task & 1.87 $\pm$ 0.09 & 209.54 $\pm$ 10.08\\
Synatra    & RTX A6000 & website & 1.25 $\pm$ 0.048 & 227.25 $\pm$ 8.73\\
\bottomrule
    \end{tabular}
    \caption{Energy cost per Token for each Split on all GPUs for the Synatra web agent}
    \label{tab:Synatra}
\end{table*}

\begin{table*}[]
    \centering
    \begin{tabular}{llrrr}
    \toprule
     Agent & GPU        & Split   &    Energy(kWh)  & Energy/Token ($10^{-9}$kWh) \\
     \midrule
Synapse    & A100-SXM4 & domain & 1.61 $\pm$ 0.05 & 234.05 $\pm$ 7.27\\
Synapse    & A100-SXM4 & task & 0.64 $\pm$ 0.014 & 215.49 $\pm$ 4.71\\
Synapse    & A100-SXM4 & website & 0.374 $\pm$ 0.011 & 193.82 $\pm$ 5.7\\
Synapse    & A100-PCIe & domain & 1.54 $\pm$ 0.055 & 223.88 $\pm$ 8.0\\
Synapse    & A100-PCIe & task & 0.59 $\pm$ 0.021 & 198.65 $\pm$ 7.07\\
Synapse    & A100-PCIe & website & 0.354 $\pm$ 0.005 & 183.46 $\pm$ 2.59\\
Synapse    & H100-SXM5 & domain & 1.17 $\pm$ 0.022 & 170.09 $\pm$ 3.2\\
Synapse    & H100-SXM5 & task & 0.458 $\pm$ 0.008 & 154.21 $\pm$ 2.69\\
Synapse    & H100-SXM5 & website & 0.272 $\pm$ 0.004 & 140.96 $\pm$ 2.07\\
Synapse    & H100-NVL & domain & 1.07 $\pm$ 0.018 & 155.55 $\pm$ 2.62\\
Synapse    & H100-NVL & task & 0.422 $\pm$ 0.004 & 142.09 $\pm$ 1.35\\
Synapse    & H100-NVL & website & 0.252 $\pm$ 0.004 & 130.6 $\pm$ 2.07\\
Synapse    & H200-SXM5 & domain & 1.13 $\pm$ 0.023 & 164.27 $\pm$ 3.34\\
Synapse    & H200-SXM5 & task & 0.438 $\pm$ 0.011 & 147.47 $\pm$ 3.7\\
Synapse    & H200-SXM5 & website & 0.262 $\pm$ 0.004 & 135.78 $\pm$ 2.07\\
Synapse    & L40S & domain & 1.56 $\pm$ 0.044 & 226.79 $\pm$ 6.4\\
Synapse    & L40S & task & 0.61 $\pm$ 0.008 & 205.38 $\pm$ 2.69\\
Synapse    & L40S & website & 0.354 $\pm$ 0.005 & 183.46 $\pm$ 2.59\\
Synapse    & RTX A6000 & domain & 2.28 $\pm$ 0.183 & 331.46 $\pm$ 26.6\\
Synapse    & RTX A6000 & task & 0.922 $\pm$ 0.105 & 310.43 $\pm$ 35.35\\
Synapse    & RTX A6000 & website & 0.5 $\pm$ 0.047 & 259.12 $\pm$ 24.36\\
\bottomrule
    \end{tabular}
    \caption{Energy cost per Token for each Split on all GPUs for the Synapse web agent}
    \label{tab:synapse}
\end{table*}
\newpage

\subsection{Carbon Dioxide Emissions Equivalents}
To provide a more complete overview of the $CO_2$ emission of each web agent, based on their energy consumption, we provide $CO_2$ equivalents for all conducted benchmark tests. \Cref{tab:co2} expands our $CO_2$ emission table in the paper, which only included the equivalents for the H100-NVL GPU. For comparison, our theoretical estimations for MindAct and LASER~\cite{ma2023laser} are also attached, showcasing that even the most inefficient GPU coupled with the most inefficient open-source web agent in our lineup uses far less energy than our energy cost estimation for LASER.

\begin{table*}[h]
\centering
    \begin{tabular}{lllrrr}
    \toprule
Agent & GPU & Energy (kWh) & Norway (g$CO_2$) & US (g$CO_2$) & Australia (g$CO_2$)\\
\midrule
AutoWebGLM & A100-SXM4 & 0.46 & 9 & 208 & 368\\
AutoWebGLM & A100-PCIe & 0.48 & 9 & 217 & 384\\
AutoWebGLM & H100-SXM5 & 0.36 & 7 & 163 & 288\\
AutoWebGLM & H100-NVL & 0.33 & 6 & 149 & 264\\
AutoWebGLM & H200-SXM5 & 0.35 & 7 & 158 & 280\\
AutoWebGLM & L40S & 0.58 & 11 & 262 & 463\\
AutoWebGLM & RTX 3090 & 0.8 & 16 & 362 & 640\\
AutoWebGLM & RTX A6000 & 0.76 & 15 & 344 & 608\\
\addlinespace
MindAct    & A100-SXM4 & 1.9 & 38 & 860 & 1520\\
MindAct    & A100-PCIe & 1.39 & 27 & 629 & 1112\\
MindAct    & H100-SXM5 & 1.58 & 31 & 715 & 1264\\
MindAct    & H100-NVL & 1.22 & 24 & 552 & 976\\
MindAct    & H200-SXM5 & 1.51 & 30 & 684 & 1208\\
MindAct    & L40S & 2.01 & 40 & 910 & 1607\\
MindAct    & RTX 3090 & 1.28 & 25 & 579 & 1024\\
MindAct    & RTX A6000 & 2.57 & 51 & 1164 & 2056\\
\addlinespace
MultiUI    & A100-PCIe & 1.18 & 23 & 534 & 944\\
MultiUI    & H100-SXM5 & 0.93 & 18 & 421 & 744\\
MultiUI    & H100-NVL & 0.822 & 16 & 372 & 657\\
MultiUI    & H200-SXM5 & 0.89 & 17 & 403 & 712\\
MultiUI    & L40S & 1.52 & 30 & 688 & 1216\\
MultiUI    & RTX 3090 & 2.0 & 40 & 906 & 1600\\
MultiUI    & RTX A6000 & 2.0 & 40 & 906 & 1600\\
\addlinespace
Synapse    & A100-SXM4 & 2.62 & 52 & 1186 & 2096\\
Synapse    & A100-PCIe & 2.48 & 49 & 1123 & 1984\\
Synapse    & H100-SXM5 & 1.9 & 38 & 860 & 1520\\
Synapse    & H100-NVL & 1.74 & 34 & 788 & 1392\\
Synapse    & H200-SXM5 & 1.83 & 36 & 828 & 1464\\
Synapse    & L40S & 2.4 & 48 & 1087 & 1920\\
Synapse    & RTX A6000 & 3.7 & 74 & 1676 & 2960\\
\addlinespace
Synatra    & A100-SXM4 & 4.53 & 90 & 2052 & 3624\\
Synatra    & A100-PCIe & 4.36 & 87 & 1975 & 3488\\
Synatra    & H100-SXM5 & 3.89 & 77 & 1762 & 3112\\
Synatra    & H100-NVL & 3.31 & 66 & 1499 & 2648\\
Synatra    & H200-SXM5 & 3.74 & 74 & 1694 & 2992\\
Synatra    & L40S & 6.33 & 126 & 2867 & 5064\\
Synatra    & RTX A6000 & 8.72 & 174 & 3950 & 6976\\
\midrule
\textit{Estimation:} MindAct &&9.01&180&4081&7208\\
\textit{Estimation:} LASER &&99.21& 1984& 44942&79368\\
\bottomrule
    \end{tabular}
    \caption{The $CO_2$ consumed by running the web agents on different GPUs for the Mind2Web benchmark evaluated on the energy mix of Norway, US and Australia. Additional theoretical estimation for MindAct and LASER.}
    \label{tab:co2}
\end{table*}

\end{document}